# A Study on Lip Localization Techniques used for Lip reading from a Video

**S.D. Lalitha**
*Assistant Professor, Department of IT, R.M.K. Engineering College,
Kavaraipettai, Thiruvallur Dist., Tamil Nadu, India.
E-mail: sdl.it@rmkec.ac.in*

**K.K. Thyagharajan**
*Professor & Dean (Academic), Department of ECE, R.M.D. Engineering
College, Kavaraipettai, Thiruvallur Dist., Tamil Nadu-601206, India. E-mail:
kkthyagharajan@yahoo.com*

**Abstract**
In this paper some of the different techniques used to localize the lips from the face are discussed and compared along with its processing steps. Lip localization is the basic step needed to read the lips for extracting visual information from the video input. The techniques could be applied on asymmetric lips and also on the mouth with visible teeth, tongue & mouth with moustache. In the process of Lip reading the following steps are generally used. They are, initially locating lips in the first frame of the video input, then tracking the lips in the following frames using the resulting pixel points of initial step and at last converting the tracked lip model to its corresponding matched letter to give the visual information. A new proposal is also initiated from the discussed techniques. The lip reading is useful in Automatic Speech Recognition when the audio is absent or present low with or without noise in the communication systems. Human Computer communication also will require speech recognition.

## Introduction

Lip localization is an active undergoing research topic with lot of improvements that recovers various difficulties faced in the research. The research of this topic has started from many more years back. But still the topic is analysed with various problems and developed with improved accuracy levels for different applications. Lip localization is also known as lip segmentation. Lip localization is the first stage of Lip reading process. Lip reading is a process where visual information is extracted from the video by watching lip movements of the speaker with or without sound. For visual information extraction reliable mouth movements are required.

There are several techniques used to localize the lips. They are taking different type of inputs and providing tracked lip outputs with different accuracy levels. Some of the techniques [1] - [21] [24] [25] are applied on front view of the face from which lip is localized. Some of the techniques [22] [23] are also applied on side view of the face to localize the lips.

The methods used for extracting lips may be semi-automatic [17] or automatic [20]. The semi-automatic methods need to have manual selection of a pixel point to initiate the process in the first image or frame whereas automatic methods will not require manual selection and it will automatically start the process to localize the lips on the face.

We have three kinds of models for segmenting lips. The one among the models, is Low level or Image based [10] [11] uses mouth region of the image. In this model, to segment the lips, features of lips and skin pixels are used. This model is finding out the height and width of the lips, not the edges of the lips to locate lips.

The other model is High level or Model based [12]-[14] uses integrity constraints and pixel information to segment the lips. This model is finding the corner of the lips to detect accurate lips than Low level model.

Also we have Hybrid model [1] - [11] & [15] -[25]which is using the parameters of both the models and finding lip width, height and corners to extract the lips more accurately than Low level and High level models.

The methods used to extract lips are either using only geometrical information [1] - [3], [7] - [17], [22] & [23] or Hybrid [4]-[6]&[18]-[25] which uses geometrical and color information of the face to track the lips.

The following sections of this paper are as follows. Section II – Methodologies used for Lip Localization, discussing different methods that are studied along with their brief description, Section III – Applications , Section IV- Summary of the papers with the methods followed in a table format, Section V – A proposed method, Section VI - Conclusion and Future Work and finally References.

## Methodologies Used For Lip Localization

*A: Semi-Automatic Lip tracking using Geometric information [17]*

A semi-automatic lip tracking algorithm is proposed by N. Eveno, A. Caplier, and P-Y Coulon [17] and it is using geometric information of the mouth to track the lips from a video. There are basically two steps followed. They are,

*Step 1:* Lip contour detection by finding upper, lower and corner pixel points in the first frame of the video input. *Step 2:* Lip tracing in the following frames using the pixels of Lip contour got from first frame by adjusting the pixel points and fitting the lip model.

*1) Lip contour detection in the first frame:*

A jumping snake algorithm [17] is used here to find out upper lips boundary points in the first frame. This algorithm is derived from classical snake algorithm [1]. Classical snake algorithm has two disadvantages. They are parameter selection and initial position selection. Both are improved in the jumping snake algorithm.

Initially, a seed (a pixel point) is identified which should be on above the mouth and near the vertical symmetry axis of the mouth manually. From the pixel point selected, snake is extended till it reaches the number of points that are already predefined by M. O. Berger and R. Mohr [2] which is equal to six key points covering outer contour and two points covering inner contour.

Every time when the snake is formed, selected seed jumps to a new position to be near to the upper edge of lips. The process is continued maximum until the distance of the jump from previous seed is smaller than a threshold. When the snake is extended, also left and right pixel points are additionally added to the snake which should be at same distance from the seed at its horizontal axis.

When the snake attains its number of points, the next seed is determined. Then a new snake is extended using the same above steps. The process is stopped when the jump's amplitude is becoming lesser than one pixel around four or five jumps. Finally, eight key points covering lip contour are found.

A Model fitting is done using parametric model. Y. Tian, T. Kanade, and J. Cohn [6] have developed a model using two parabolas. T. Coianiz, L. Torresani, and B. Caprile [7] used two parabolas only for upper lip and one parabola is used for lower lip. Then B. D. Lucas and T. Kanade [8] introduced four parabolas to find the lips which are not suitable for assymetric mouth. The model used here in the method [17] has five curves to cover the lips which are more accurate and also suitable for asymmetric mouth.

Lip corners are identified after fitting a model to extract the lips more accurately.

*2) Tracing lips in the following frames:*

Here, to trace the lips in the following frames of the video the same procedure is not repeated as such in the first frame. Using the key points, we got in previous step, easily tracking is done in the next frames by finding their positions with the help of Kanade – Lucas algorithm [9].

The method is used on bearded speaker, mouth with teeth, tongue as well as on speech with noise.
But here the drawback is manually we have to select the initial seed to start the detection.

*B. (ALiFE) Automatic Lip Feature Extraction using Geometric information [20]*

ALiFe is an automatic method developed to locate the lips for automatic speech recognition. It follows three steps.

*Step 1*: Segmenting Lips in the first frame

*Step 2:* Tracing Lips in the successive frames

*Step 3:* Using the Lip detection Viseme Classification

*1) Lip Segmentation:* Lip segmentation follows two steps. *a. Detecting Lip Contour:*
Lip contour algorithms [16] [17] are applied to segment the lips on the mouth. Initially a model of snake is placed to detect the actual lips in the mouth region. Then the model slowly shrinks according to the energies accompanied to the snake which is closing towards the actual lips. The energies associated are internal energy due to the contour of the shape, external energy imposed by the image and supplementary constraints imposed by the user. The snake extension is stopped when total energy becomes minimal according to energy minimization principle or to a maximum number of iterations.

*b. POI (Points Of Interest) Identification:*
Horizontal and vertical points of the contour are identified after finding lip contour. Horizontal points are lip edges which will be found by maximum extension in the horizontal axis of the contour. Vertical points are the height of the lips which is used to cover the boundary of the lips and they are maximum extension of vertical axis.

*2) Tracking lips:*
Here the detected POIs from the first frame are traced in the following frames block by block. This method is called as Template matching method. In this method, spatial temporal indices of the input are considered. It searches a block which is similar to the first frame that covers the POI in the following frames in a gray level image.
It uses two steps. They are as follows.
   a. In various directions POI will be tracked to detect the POI movements in the following frames, and candidate points are derived.
   b. Among the derived candidate points, vote method is used to find the nearest POI in the following frames.

*3) Extracting Visual Information:*
After segmenting the lips based on the model of the lips, each frame will give a letter to generate visual information.

The method is automatic method which will not require manual initialization to start the detection. The method uses geometric information and it will also be handling asymmetric mouth with visible teeth and tongue.

## C. Automatic Hybrid Method using Geometric and Color Information [21]

A hybrid method is proposed [21] which is also an automatic method. It uses both geometric and color information for tracking POIs on mouth to locate lips. This method follows two steps. They are,

*Step 1:* Identifying POIs to segment the Lip contour using Color information and then also by Geometric information in the first frame of video.

*Step 2:* Lip tracing in the successive frames of video.

*1) Identifying POIs:*
   It uses two steps as follows.

a. *Detecting Mouth Corners using Color information:* Mouth color is separated from skin color using HSV color model and rg chromaticity [18] to extract mouth region from face. For mouth localization, Yasuyuki Nakata and Moritoshi Ando [19] method is used where color distribution based on relationship of normalized RGB Values of each organ of face.
So the first frame is specified in RnGnBn color system.

Then binarization is applied on the image using a binary threshold based on Rn value which is more on lip area. An oily filter is applied on the result of binarization to get only lip pixels excluding false red skin pixels. Saturation component is used to localize the mouth corners from the mouth region.

b. *Detecting POIs using Geometrical information:*
Using the mouth corners detected from the previous step, the lip model specified by Alan Wee-Chung Liewa, Shu Hung Leungb and Wing Hong Laua [25] is used to detect POIs to extract lip contour.

*2) Tracing Lips in the following frames:*
The lip tracking in the successive frames are done referring the method used in [20].

The hybrid solution is useful to tolerate noise and artefacts in the image more than the methods using only geometrical or color model.

## Applications

There are many fields where Lip localization is playing important role. It is used by deaf and dumb people who are having hearing impairments to understand the speaker words. It will also used for understanding the speech when the voice is low with voice.

It could be used to translate an unknown language to a known language. Automatic speech recognition is used in video conferences, multimedia systems and low communication systems. Also, Lip localization is used to identify the speaker.

## Summary

Here some of the papers used for extracting lips are listed with the brief concept and the technique applied in the concept.

**Table 1:** Summary Table

| Concept | Paper - Year | Method |
| --- | --- | --- |
| Locating lips using hue color then edge of the lip is identified by points on edge. | Using Lip Features for Multimodal Speaker Verification- 2001. | Automatic – Hybrid Features |
| Lip contour is detected using red hue values and then using geometric features, lip contour is located. | Feature extraction & lips posture detection oriented to the treatment of CLP children – 2006. | Automatic – Hybrid Features |
| Lip contour identified then points on contour identified using Geometric features. | Audio - Visual Speech Recognition using Lip information extracted from side-face Images, 2007. | Automatic - Geometric Features |
| On the identified mouth region, using color model lip corners are identified then height and width of the lips are identified using geometric features. | Combining Edge Detection and Region segmentation for Lip Contour Extraction, 2008. | Automatic – Hybrid Features |
| Inner height, width and outer height & width are identified using geometric features. | A novel Approach Integrating Geometric and Gabor Wavelet Approaches to improvise Visual Lip reading, 2010. | Automatic – Geometric Features |
| Lip area is identified by pixel intensity then lip edge points are extracted. | Lip contour detection techniques based on front view of face – 2011. | Automatic – Geometric Features |
| Lip extraction using Geometry information- height, width and ratio of height and width features. | Geometry based lip reading system using Multi Dimension Dynamic Time Warping – 2012. | Automatic -- Geometric Features |
| From mouth corners then five key points on the lips are identified using geometric features. | Impact of the Lips for Biometrics, 2012. | Automatic – Geometric Features |
| Lip area is extracted using color model then using geometric features, lip is located | An algorithm of lips secondary positioning and feature extraction based on YCbCr color space - 2015 | Automatic – Hybrid Features |
| Lip corners are identified using pixel model then it identifies lip contour using improved jumping snake algorithm | Automatic Lip Contour Extraction Using Both pixel-based and parametric models | Automatic – Geometric Features |

## A Proposed Method

We have a new proposal for lip recognition from the above methods discussed.

*Steps:*
a. Finding differences between the frames of the video input, using the method AMI (Accumulated Motion Image) [26] to find out the portion of the frame where the movements are happening from which the moved facial organ's (lips & eyes mainly) pixels will be extracted.

b. We can now apply a color model on the result of AMI to extract only the lip pixels where red pixels are dominant from the other pixels moved (eyes).
c. Then we can apply LBP (Local Binary Pattern) [27] to find out edge information of lips.
d. Using Geometric features, other pixel points covering lip contour (Outer height, width, inner height & width) can be detected to extract lip contour completely along with edge information.
e. Using a trained data set, we could generate visual information after extracting lips from the face input.

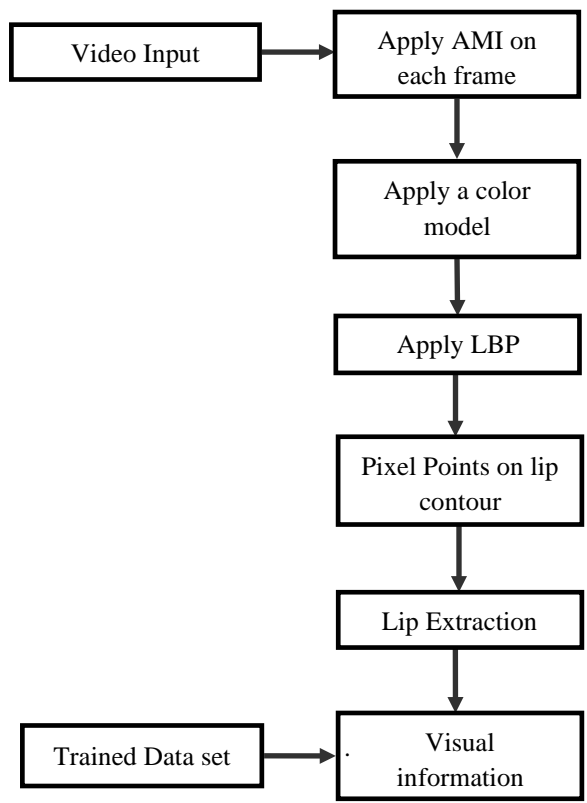

**Figure 1:** Overview of a Proposal

**Conclusion and Future Work**

In this paper, we have presented a study on some of the lip localization techniques. Among the techniques, one of the techniques is semi-automatic. It uses geometric information and a manual selection of a pixel point is required for initializing the lip contour detection.

Other technique is automatic method using geometric information of image and not using manual selection for initialization.

The third method in the paper, discussed about hybrid which is automatic as well as using geometric and color information for detecting lips.

Then a summary table is drawn with list of some papers using either geometric or hybrid method for lip detection.

Finally we have proposed a method to identify lips and to generate visual information with our suggested steps.

Although the paper discusses some of the methods that are already available, we want to study some more different methods. We also planned to workout our proposed method with a trained dataset in our future.

# References

[1] M. Kass, A. Witkin, and D. Terzopoulos, Jan. 1988, "Snakes: Active contour models," Int. J. Comput. Vis., vol. 1, no. 4, pp. 321–331.
[2] M. O. Berger and R. Mohr, 1990, "Toward autonomy in active contour models," in Proc. ICPR, pp. 847–851.
[3] N. Eveno, A. Caplier, and P. Y. Coulon, 2002, "Key points-based segmentation of lips," in Proc. ICME, pp. 125–128.
[4] A. Hulbert and T. Poggio, 1998, "Synthesizing a color algorithm from examples," Science, vol. 239, pp. 482–485.
[5] N. Eveno, A. Caplier, and P. Y. Coulon, 2001, "A new color transformation for lips segmentation," in Proc. MMSP, pp. 3–8.
[6] Y. Tian, T. Kanade, and J. Cohn, 2000, "Robust lip tracking by combining shape, color and motion," in Proc. ACCV, pp. 1040–1045.
[7] T. Coianiz, L. Torresani, and B. Caprile, 1995, "2D deformable models for visual speech analysis," NATO Advanced Study Institute: Speech reading by Man and Machine, pp. 391–398.
[8] M. E. Hennecke, K. V. Prasad, and D. G. Stork, 1994, "Using deformable templates to infer visual speech dynamics," in Proc. 28th Annu. Asilomar Conf. Signals, Systems, and Computers, pp. 578– 582.
[9] B. D. Lucas and T. Kanade, 1981, "An iterative image registration technique with an application to stereo vision," in Proc. IJCAI'81, Vancouver, BC, Canada, pp. 674–679.
[10] I. Matthews, J. Andrew Bangham, and Stephen J. Cox. , October 1996, "Audiovisual speech recognition using multiscale nonlinear image decomposition". Proc. 4th ICSLP, volume1, page 38-41, Philadelphia, PA, USA.
[11] U. Meier, R. Stiefelhagen, J. Yang et A. Waibe, Jan 1999, "Towards unrestricted lip reading". Proc 2nd International conference on multimodal Interfaces (ICMI), Hong-kong.
[12] K. Prasad, D. Stork, and G. Wolff, September 1993, "Preprocessing video images for neural learning of lipreading," Technical Report CRCTR- 9326, Ricoh California Research Center.
[13] R. Rao, and R. Mersereau, 1995, "On merging hidden Markov models with deformable templates," ICIP 95, Washington D.C.
[14] P. Delmas, N. Eveno, and M. Lievin, 2002, "Toward robust lip tracking," in Proc. ICPR, pp. 528–531. [15] Sunil S. Morade and B. Suprava Patnaik, "April 2013Automatic Lip Tracking and Extraction of Lip Geometric Features for Lip Reading", International Journal of Machine Learning and Computing, Vol 3, No.2.
[16] N. Eveno, A. Caplier, and P. Y. Coulon, 2003, "Jumping snake and parametric model for lip segmentation," in Proc. ICIP, Barcelona, Spain, pp. 867–870.
[17] N. Eveno, A. Caplier, and P-Y Coulon, May 2004, "Accurate and Quasi-Automatic Lip Tracking", IEEE Transaction on circuits and video technology. [18] Miyawaki T, Ishihashi I, Kishino F, 1989," Region separation in color images using color information", Tech Rep IEICE; IE89-50.
[19] Nakata Y, Ando M, 2004, "Lipreading Method Using Color Extraction Method and Eigenspace Technique", Systems and Computers in Japan, Vol. 35, No. 3.
[20] Salah Werda, Walid Mahdi and Abdelmajid Ben Hamadou, April 2007, "Lip Localization and Viseme Classification for Visual Speech Recognition", International Journal of Computing & Information Sciences Vol.5, No.1, Pg. 62-75.
[21] Salah Werda, Walid Mahdi and Abdelmajid Ben Hamadou, April 12-14, "Automatic Hybrid Approach for Lip POI Localization: Application for Lip-reading System", ICTA'07, Hammamet, Tunisia.
[22] Patrick Lucey, Gerasimos Potamianos, Sridha Sridharan, "An extended Pose –Invariant Lip reading System", Speech, Audio, Image and Video Technology Laboratory, Queensland University of Technology, Brisbane, Australia.
[23] Kji lwano, Tomoaki Yosh inaga, Santoshi Tamura and Sadaoki Furui, January 2007,  "Audio-Visual speech Recognition Using Lip information Extracted from side-face Images", EURASIP Journal and Audio,Speech, and Music Processing.
[24] Vicente P. Minotto, Carlos B.o. Lopes,Jacob Scharcanski,  Claudi R. Jung and Bowon Lee,  February 2013, "Audiovisual Voice activity Detection Based on Microphone Arrays and Color Information", IEEE Journal of selected topics in Signal Processing.
[25] Alan Wee-Chung Liewa, Shu Hung Leungb, Wing Hong Laua, Nov 2002, "Lip contour extraction from color images using a deformable model", The Journal of Pattern Recognition Society.
[26] Thanikachalam V, Thyagharajan K K, 2015, "Human Action Recognition Using Motion History Image and Correlation Filter", Special Issue of International Journal of Applied Engineering Research, Vol. 10, No.34, pp. 27361-27363
[27] Thanikachalam V, Thyagharajan K K, 2012, "Human Action Recognition Using Accumulated Motion and Gradient of Motion from Video" Proceedings of the Third International Conference on Computing Communication and Networking Technologies ICCCNT 2012, IEEE Explore.